# Multi-Label MRF Optimization via Least Squares $s-t$ Cuts

Ghassan Hamarneh
Simon Fraser University
8888 University Dr., Burnaby, BC

hamarneh@cs.sfu.ca

October 25, 2018

**Abstract**

There are many applications of graph cuts in computer vision, *e.g.* segmentation. We present a novel method to reformulate the NP-hard, k-way graph partitioning problem as an approximate minimal $s-t$ graph cut problem, for which a globally optimal solution is found in polynomial time. Each non-terminal vertex in the original graph is replaced by a set of $ceil(log_2(k))$ new vertices. The original graph edges are replaced by new edges connecting the new vertices to each other and to only two, source $s$ and sink $t$, terminal nodes. The weights of the new edges are obtained using a novel least squares solution approximating the constraints of the initial k-way setup. The minimal $s-t$ cut labels each new vertex with a binary ($s$ vs $t$) "Gray" encoding, which is then decoded into a decimal label number that assigns each of the original vertices to one of k classes. We analyze the properties of the approximation and present quantitative as well as qualitative segmentation results.

## 1 Introduction

Many computer vision problems can be formulated as graph labeling problems, *e.g.* segmentation [9], denoising [7], registration [33, 42], point correspondence from stereo pairs [3], and shape matching [38]. The general problem is to assign one label, out of a finite set of labels, to each vertex of the graph in some optimal and meaningful way. In some cases, this may be formulated as a graph cut problem, where the task is to separate the vertices into a number of groups with a common label assigned to all vertices in each group.

## 1.1 Graph cuts without label interaction

The multi-way or *k*-way cut problem is defined as follows. Given a graph $G(V,E)$ with vertices $v_j \in V$ and edges $e_{v_i,v_j} = e_{ij} \in E \subseteq V \times V$ with positive edge weights $w(e_{ij})$ or $w_{ij}$, $w_{ij} = w_{ji}$ (*i.e.* undirected graph), find an optimal *k*-way cut (or *k*-cut) $C^* \subset E$ with minimal cost $|C^*| = argmin_C |C|$, where $|C| = \sum_{e_{ij} \in C} w_{ij}$, such that restricting the edges to $E \setminus C$ breaks the graph into *k* disconnected graphs dividing the vertices into *k* groups, with all vertices in one group sharing one of the labels $\mathscr{L} = \{l_0, l_1, ..., l_{k-1}\}$. We denote the number of vertices and edges in the graph as $|V|$ (*i.e.* $V = \{v_1, \cdots, v_{|V|}\}$) and $|E|$, respectively.

This *k*-cut formulation assumes that all the semantics about the computer vision problem at hand can be encoded into the graph edge weights. For this class of cut problems the discrete optimization and graph theory communities has made notable progress. For $k = 2$, the problem reduces to partitioning $V$ into 2 sets (binary labeling). The min-cut max-flow theorem equates the minimal cut in this case ($k = 2$) to the maximal network flow from one of the terminal vertices, the source *s*, to the other, the sink *t*. For $k \geq 3$, the general multi-way cut problem is known to be NP-hard. However, for the special case of planar graphs, a solution can be found in polynomial time [45, 10]. Finding the minimal cut that separates the source from the sink for $k = 2$ can always be found in polynomial-time, *e.g.* using the Ford-Fulkerson [16], Edmunds-Karp [14], or Goldberg-Tarjan [18], even for non planar graphs. When $k = 2$ and there are constraints on the size of the partitioned sets then the problem is NP-hard. Several non-global solutions have been proposed to the multi-way cut ($k \geq 3$) problem. Dahlhaus *et al.* proposed an algorithm based on the union of individual 2-way cuts that is guaranteed a cut cost with a factor of $2 - 2/k$ of the optimal cut cost [13]. Through a linear programming relaxation algorithm, Calinescu *et al.* improve on the approximation ratio down to at most $1.5 - 2/k$ [12]. For $k = 3$, the $7/6 (= 1.5 - 2/3)$ factor of [12] is further improved by Karger *et al.* to $12/11$ using a geometric embedding approach [26]. Zhao *et al.* approximate the *k*-way cut via a set of minimum 3-way cuts with an approximation ratios of $2 - 3/k$ for an odd *k* and $2 - (3k-4)/(k^2-k)$ for an even *k* [31]. In [19], Goldschmidt and Hochbaum show that the NP-completeness of the specified vertices problem does not imply NP-completeness of the *k*-cut problem without specified fixed terminal vertices.

## 1.2 Markov random fields

In computer vision applications, it is sometimes difficult, if not impossible, to encode the semantics of the computer vision problem into the topology and edge weights of a graph, and proceed by applying a graph cut method. On the contrary, it is often desirable that the label assignment to a vertex in the graph be influenced by the labels assigned to other vertices. This means that the labeling resulting from the cut has an impact on the cost of the cut itself (*i.e.* does not depend only on the edge weights severed). In Markov random fields (MRF) [17], modeling a vertex label's dependence on the labels of all other vertices is captured by the dependence only on the labels of the immediate neighbors. In image segmentation, for example, each pixel is represented by a vertex in a graph and the graph edges capture the neighborhood relationship be-tween pixels (*e.g.* via 4- or 8-connectivity in 2D). It is usually desirable that the graph partitioning and, hence, the induced labeling satisfy two possibly conflicting criteria: (i) a pixel is labelled according to the data (*e.g.* the image) value at that particular vertex (*e.g.* pixel) only; and (ii) neighboring vertices (e.g. pixels) are assigned identical or similar labels. In general, this latter criterion regularizes the solution and makes it more robust to noise.

MRF theory can be adopted to formulate this desired behavior as an objective function to be minimized, using discrete optimization methods, with respect to the different possible vertex labels over the whole graph. Given a graph $G(V,E)$ the MRF energy function can be written as

$$\xi(l) = \sum_{v_i \in V} D_i(l_i) + \lambda \sum_{(v_i,v_j) \in E} V_{ij}(l_i, l_j, d_i, d_j) \quad (1)$$

where $v_i$ and $v_j$ are vertices (*e.g.* corresponding to two pixels $p$ and $q$) with data values $d_i$ and $d_j$ (*e.g.* the image values $I(p)$ and $I(q)$) and with labels $l_i$ and $l_j$, respectively. $D_i(l_i)$ is the data term (or image fidelity) measuring the penalty of labeling $v_i$ with a specific label $l_i$, disregarding the labels or data values of any of the other (neighbors or elsewhere) vertices, and $V_{ij}$ is the vertex interaction term that penalizes certain label configurations of neighboring vertices $v_i$ and $v_j$, *i.e.* the penalty of assigning label $l_i$ to $v_i$ and $l_j$ to $v_j$. $\lambda$ controls the relative importance of the two terms. $V_{ij}$ can be seen as a metric on the space of labels $V_{ij} = V_{ij}(l_i, l_j)$ (also called the *prior*) or may be chosen to depend on the underlying data $V_{ij} = V_{ij}(d_i, d_j)$, or both $V_{ij}(l_i, l_j, d_i, d_j)$, *e.g.*

$$V_{ij}(l_i, l_j, d_i, d_j) = V_{ij}^l(l_i, l_j) V_{ij}^d(d_i, d_j) \quad (2)$$

where superscripts $l$ and $d$ denote label and data interaction penalties, respectively. Various label interaction penalties have been proposed, including linear: $V_{ij}^l = |l_i - l_j|$, quadratic penalty: $(l_i - l_j)^2$, truncated versions thereof: $\min\{T, |l_i - l_j|\}$ or $\min\{T, (l_i - l_j)^2\}$, with threshold $T$, or Kronecker delta $\delta_{l_i \neq l_j}$ [5]. Various spatially-varying penalties depending on the underlying image data have also been proposed, *e.g.* Gaussian $V_{ij}^d = e^{-\beta(d_i-d_j)^2}$ or reciprocal $\frac{1}{1+\beta(d_i-d_j)^2}$ [21].

## 1.3 MRF for computer vision

In computer vision applications, *e.g.* segmenting an image into $k$ regions, an image with $P$ pixels is typically modeled as a graph with $P$ vertices, one for each pixel (*i.e.* each vertex is mapped to a location in $\mathbb{Z}^d$, where $d$ is the dimensionality of the image). To encode the data term penalty $D_i(l_i)$, the graph is typically augmented with $k$ new terminal vertices $\{t_j\}_{j=1}^k$; each representing one of the $k$ labels (Figure 2a). The edge weight connecting a non-terminal vertex $v_i$ (representing pixel $p$) to a terminal vertex $t_j$ (representing label $l_j$) is set inversely proportional to $D_i(l_j)$; the higher the penalty of labeling $v_i$ with $l_j$ the smaller the edge weight and hence the more likely the edge will be severed, *i.e.*

$$w_{v_i,t_j} \propto 1/D_i(l_j); \forall v_i \in V, \forall t_j \in \{t_j\}_{j=1}^k. \quad (3)$$

Setting the edge weights according to any vertex interaction penalty $V_{ij}(l_i,l_j,d_i,d_j)$ is not straightforward. In the special case when $V_{ij}(l_i,l_j,d_i,d_j) = V_{ij}^d(d_i,d_j)$, *i.e.* independent of the labels $l_i$ and $l_j$, then typically $V_{ij}$ is encoded through the graph edge weights connecting vertices representing neighboring pixels. The higher the edge weight the less likely the edge will be severed and hence the more likely the two pixels will be assigned the same label. This motivates setting the edge weights proportional to $V_{ij}^d$, *i.e.*

$$w_{v_i,v_j} \propto V_{ij}(d_i,d_j); \forall e_{ij} \in E. \qquad (4)$$

This approach, however, discourages (or encourages) cutting the edge between neighboring vertices and hence assigning the same (or different) labels to the vertices without any regard to what these same (or different) labels are . Clearly, (4) is not flexible enough to encode more elaborate label interactions (since essentially $V_{ij}^l(l_i,l_j) = constant$). In fact, this issue is at the heart of the challenging multi-label MRF optimization problem: developing globally (or close to global) optimal algorithms for any interaction penalty.

Greig *et al.* presented one of the earliest works (1989) on combinatorial optimization approaches to a computer vision problem [22]. They constructed a two-terminal graph, whose minimal cut gives a globally optimal binary vector used for restoring binary images. In earlier works, iterative algorithms, such as simulated annealing where employed to solve MRF problems. Later (1993), Wu and Leahy applied a graph theoretic approach to data clustering for image segmentation [44]. The 1997 work of Shi and Malik on normalized cuts [39, 40] sparked large interest in graph-based image partitioning. In [39], the cost of a partition is defined as the total edge weight connecting two partitions as a fraction of the total edge connections to all the nodes in the graph, which is written as a Rayleigh quotient whose global minimum is obtained as the solution of a generalized eigen-system. In 1998, Roy and Cox [36] re-formulated the multi-camera correspondence as a max-flow min-cut problem. Boykov and Jolly applied the min-cut max-flow graph cut algorithm to find the globally optimal binary segmentation [6].

As mentioned earlier, in the multi-label problem ($k \geq 3$) the global minima is generally not attainable in polynomial time. In the special case of convex label interaction penalty, also known as *convex prior*, Ishikawa proposed a method that achieves the global energy minimizer [25, 24]. Ishikawa's convex prior condition is given by:

$$2V_{ij}^l(l_i - l_j) \geq V_{ij}^l(l_i - l_j + 1) + V_{ij}^l(l_i - l_j - 1) \qquad (5)$$

This convex definition was later generalized in [37]. However, with convex priors, *e.g.* quadratic $V_{ij}^l = (l_i - l_j)^2$, the penalty for assigning different labels to neighboring pixels can become excessively large, which in turn over-smoothes the label field because several small changes in the label can yield a lower cost than a single sudden change. This encourages pixels at opposite sides of an interface between two different regions be assigned the same or similar labels albeit ideally they shouldn't. This motivates the introduction of non-convex priors, typically achieved by truncating the penalty (*e.g.* the truncated quadratic $\min\{T,(l_i-l_j)^2\}$ or the Pott's model), to allow for discontinuities in the label field at the cost of no longer guaranteeing a globally optimal energy minimizer.

This tradeoff, either a guaranteed global minima of convex prior or a *discontinuity-preserving* prior whose global minima cannot be achieved, has sparked a strong interest within the computer vision community in improving the state-of-the-art of optimizing multi-label problems with non-convex priors. In their seminal work, Geman and Geman applied simulated annealing based optimization [17]. In [2], Iterated Conditional Modes was proposed. In [39], to segment multiple regions, a recursive sub-optimal approach is used, which entails deciding if the current partition should be further subdivided and repartitioning if necessary. In a somewhat reverse approach, Felzenszwalb and Huttenlochers algorithm assigns a different label to each vertex, then similar pixels are merged using a greedy decision approach [15]. Boykov *et al.* proposed two algorithms that rely on an initial labeling and an iterative application of binary graph cuts. At each iteration, an optimal *range move* is performed to either expand ($\alpha$-expansion) or swap labels ($\alpha - \beta$-swap) [8, 9]. Although convergence and error bounds are guaranteed, the initial labeling may influence the result of the algorithm. In [43], Veksler proposes a new type of range moves that act on a larger set of labels than those in [9]. The LogCut [32] is another iterative range move based algorithm that applies the binary graph cut at successive bit-levels of binary encodings of the integer labels (from most significant to least significant) rather than once for each possible value of the labels. In [27], the image is partitioned into two regions by computing a minimum cut with a swap move of binary labels and then the same procedure is recursively applied to each region to obtain new regions until some stopping condition is met. Recently, Szeliski *et al.* presented a study comparing energy minimization methods for MRF[41].

The aforementioned range move type of approaches are regarded as the state-of-the-art in solving multi-label assignment problems in the computer vision community. It is important to note, however, that the $\alpha - \beta$ swap algorithm can only be applied in the cases when $V_{ij}^l$ is semi-metric [9], *i.e.* satisfying both conditions

$$V_{ij}^l(\alpha,\beta) = 0 \Leftrightarrow \alpha = \beta \tag{6}$$

$$V_{ij}^l(\alpha,\beta) = V_{ij}^l(\beta,\alpha) \geq 0 \tag{7}$$

On the other hand, $\alpha$-expansion is even more restricted and can only be applied when $V_{ij}^l$ is metric [9], *i.e.*, in addition to the two above conditions, the following triangular inequality must also hold

$$V_{ij}^l(\alpha,\beta) \geq V_{ij}^l(\alpha,\gamma) + V_{ij}^l(\gamma,\beta) \tag{8}$$

These label-interaction restrictions (convex, semi-metric, metric) limit the applications of graph cuts algorithms, since the semantics of the computer vision problem can not always be easily formulated to abide by these restrictions.

More recent approaches to solving the multi-label MRF optimization have been proposed based on linear programming relaxation using primal-dual [30], message passing [29], and partial optimality [28].

### 1.4 Contributions

In this work, we propose a novel method to convert the multi-label MRF optimization problem to a binary labeling of a new graph with a specific topology. The error in the new edge weights is minimized using least-squares (LS). The resulting binary labeling is solved via a single application of $s-t$ cut, *i.e.* the solution is non-iterative and does not depend on any initializations. Once the binary labeling is obtained, it is directly decoded back to the desired non-binary labeling. The method accommodates any label or data interaction penalties, *i.e.* any $V_{ij}(l_i,l_j,d_i,d_j)$, *e.g.* non-convex or non-metric priors or spatially varying penalties. Further, besides its optimality features, LS enables offline pre-computation of pseudo-inverse matrices that can be used for different graphs.

## 2 Method

### 2.1 Reformulating multi-label MRF as $s-t$ cut

Given a graph $G(V,E)$[1], the objective is to label each vertex $v_i$ with a label $l_i \in \mathscr{L}_k = \{l_0,l_1,...,l_{k-1}\}$. The key idea of our method is: Rather than labeling $v_i$ with $l_i \in \mathscr{L}_k$, we replace the vertex $v_i$ with $b$ vertices $v_{ij}, j \in \{1,2,\cdots,b\}$, and label each $v_{ij}$ with a binary label $l_{ij} \in \mathscr{L}_2 = \{0,1\}$. Assigning label $l_i$ to vertex $v_i$ in $G(V,E)$ entails assigning a corresponding sequence of binary labels $(l_{ij})_{j=1}^b$ to $(v_{ij})_{j=1}^b$. We distinguish between the decimal (base 10) and binary (base 2) encoding of the labels using the notation $(l_i)_{10}$ and $(l_i)_2 = (l_{i1},l_{i2},\cdots,l_{ib})_2$, respectively, with $l_i \in \mathscr{L}_k$ and $l_{ij} \in \mathscr{L}_2$. Consequently, the original graph $G(V,E)$ is transformed into $G_2(V_2,E_2)$, where subscript 2 denotes the binary representation (or binary encoding). $V_2$ is given as

$$V_2 = \left\{\{v_{ij}\}_{i=1}^{|V|}\right\}_{j=1}^b \cup \{s,t\}. \qquad (9)$$

with $|V_2| = b|V|+2$, *i.e.* $b$ vertices in $V_2$ for each vertex in $V$ and two terminal, source and sink, vertices $\{s,t\}$. $b$ must be chosen large enough such that the binary labeling of the $b$ vertices $v_{ij}$ can be decoded back into a label $l_i \in \mathscr{L}_k$ for $v_i$. In other words, $b$ is the number of *bits* needed to encode $k$ labels. Therefore, we must have $2^b \geq k$, or

$$b = ceil(log_2(k)). \qquad (10)$$

Each edge in $E_2$ is either a terminal link (t-link), a neighborhood links (n-links), or an intra-link, *i.e.*

$$E_2 = E_2^{tlinks} \cup E_2^{nlinks} \cup E_2^{intra}. \qquad (11)$$

where (Figure 1)

$$E_2^{tlinks} = E_2^t \cup E_2^s. \qquad (12)$$

$$E_2^{nlinks} = E_2^{ns} \cup E_2^{nf}. \qquad (13)$$

Edges in $E_2^t$ and $E_2^s$ connect each vertex in $V_2$ to terminals $t$ and $s$, respectively, therefore $|E_2^t| = |E_2^s| = |V_2|$. With $b$ vertices in $V_2$ replacing each vertex in $V$, up to $b^2$ unique edges can connect vertices in $V_2$ that correspond to neighboring vertices $i$ and

---
[1] There are no terminal vertices in this original graphs. Nevertheless, we draw terminal vertices in Figure 2(a,d) to clarify that in a multi-way cut each vertex is assigned a label.

$j$ in $V$. $E_2^{nlinks}$ contain all these $b^2$ edges for all pairs of neighboring vertices, i.e. $|E_2^{nlinks}| = b^2|E|$. We distinguish between two types of n-links: $E_2^{ns}$ is limited to the sparse set of $b$ edges that connect corresponding vertices $(v_{im}, v_{jm})$ (note same $m$ in both vertices), i.e. $|E_2^{ns}| = b|E|$, whereas $E_2^{nf}$ contains all the remaining edges that connect non-corresponding vertices $(v_{im}, v_{jn}), m \neq n$, i.e. $|E_2^{nf}| = (b^2 - b)|E|$. $E_2^{intra}$ includes edges that connect pairs of vertices $(v_{im}, v_{in})$ (note same $i$) in $V_2$ that are among the set of vertices representing a single vertex in $V$, yielding $|E_2^{intra}| = \binom{b}{2}|V|$. Formally,

$$\begin{aligned} E_2^t &= \{e_{v_{ij}t}; \forall v_{ij} \in V_2\} \\ E_2^s &= \{e_{v_{ij}s}; \forall v_{ij} \in V_2\} \\ E_2^{ns} &= \{e_{im,jm}; \forall e_{ij} \in E\} \\ E_2^{nf} &= \{e_{im,jn}; \forall e_{ij} \in E, m \neq n\} \\ E_2^{intra} &= \{e_{im,in}; \forall v_{ij} \in V_2, m \neq n\} \end{aligned} \quad (14)$$

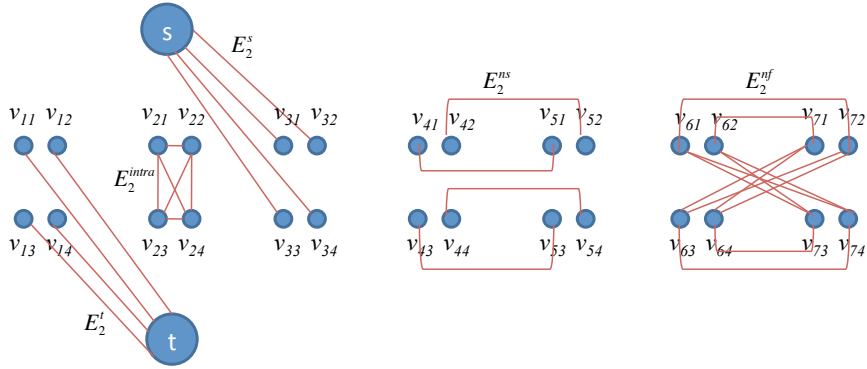

Figure 1: Edge types in the $s-t$ cut formulation. Shown are seven groups of vertex quadruplets, $b=4$, and only sample edges of $E_2^t, E_2^s, E_2^{ns}, E_2^{nf}$, and $E_2^{intra}$.

Following an $s-t$ cut on $G_2$, vertices $v_{ij}$ that remain connected to $s$ are assigned label 0, and the rest that are connected to $t$ are assigned label 1 (we could swap 0 and 1 without loss of generality). The string of $b$ binary labels $l_{ij} \in \mathscr{L}_2$ assigned to $v_{ij}$ are then decoded back into a decimal number indicating the label $l_i \in \mathscr{L}_k$ assigned to $v_i$ (Figure 2).

It is important to set the edge weights in $E_2$ in such a way that decoding the binary labels resulting from the $s-t$ cut of $G_2$ will result in optimal (or as optimal as possible) labels for the original multi-label problem. We do not expect to optimally solve the multi-label problem this way, but rather to provide an approximate solution. The second key idea of our method is: Derive a system of linear equations capturing the relation between the original multi-label MRF penalties and the $s-t$ cut cost incurred when generating different label configurations, and then calculate the weights of $E_2$ as the LS error solution to these equations. In the next sections, we show how we choose the edge weights of $E_2$ in a minimum LS error formulation.

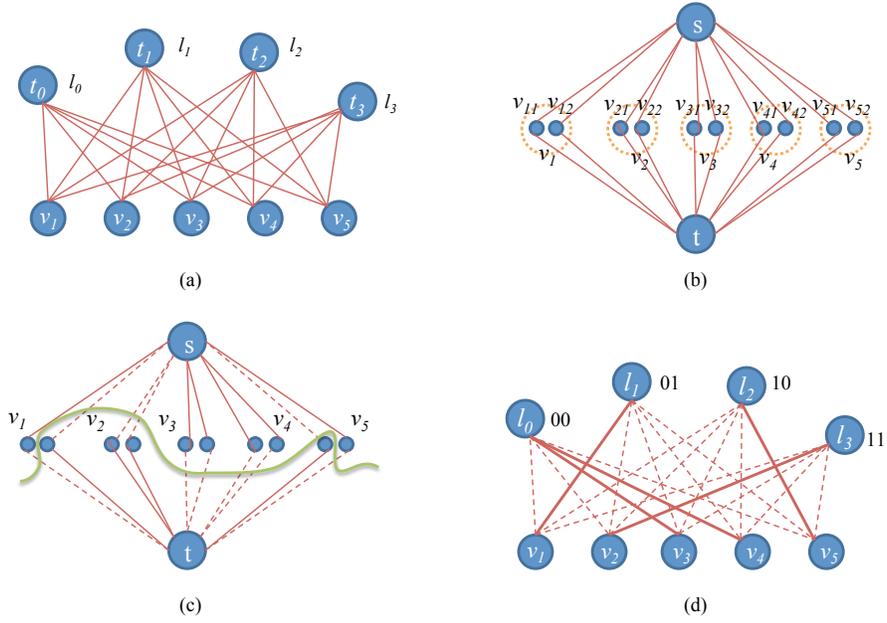

Figure 2: Reformulating the multi-label problem as a single $s-t$ cut. (a) A multi-label problem ($k$-way cut) of labeling vertices $\{v_i\}_{i=1}^{5}$ with labels $\{l_j\}_{j=1}^{4}$ (only $E_2^{tlinks}$ are shown). (b) New graph with 2 terminal nodes $\{s,t\}$, $b=2$ new vertices ($v_{i1}$ and $v_{i2}$ inside the dashed circles) replacing each $v_i$ in (a), and 2 terminal edges for each $v_{ij}$. (c) An $s-t$ cut on (b). (d) Labeling $v_i$ in (a) is based on the $s-t$ cut in (c). (d) Pairs of $(v_{i1}, v_{i2})$ assigned to $(s,s)$ are labeled with binary string 00, $(s,t)$ with 01, $(t,s)$ with 10, and $(t,t)$ with 11. The binary encodings 00, 01, 10, or 11 in turn reflect the original 4 labels.

## 2.2 Data term penalty: Severing t-link, intra-links

In the proposed binary formulation, the data term penalty $D_i(l_i)$ in (1) equals the cost of assigning label $l_i$ to vertex $v_i$ in $G(V,E)$, which entails assigning a corresponding sequence of binary labels $(l_{ij})_{j=1}^{b}$ to $(v_{ij})_{j=1}^{b}$ in $G_2(V_2, E_2)$. To assign $(l_i)_2$ to a string of $b$ vertices, appropriate terminal links must be cut. To assign a 0 (resp. 1) label to $v_{ij}$ the edge connecting $v_{ij}$ to the terminal t (resp. s) must be severed (Figure 3). Therefore, the local (corresponding only to labeling $v_i$) cost of severing t-links in $G_2$ to assign $l_i$ to vertex $v_i$ in $G$ can be calculated as

$$D_i^{tlinks}(l_i) = \sum_{j=1}^{b} l_{ij} w_{v_{ij},s} + \bar{l}_{ij} w_{v_{ij},t} \tag{15}$$

where $\bar{l}_{ij}$ denotes the unary complement (NOT) of $l_{ij}$, $w_{v_{ij},s} = w(e_{v_{ij},s})$ is the weight of the edge connecting $v_{ij}$ to $s$, and, similarly, $w_{v_{ij},t} = w(e_{v_{ij},t})$, with $e_{v_{ij},s} \in E_2^s$ and $e_{v_{ij},t} \in E_2^t$.

The $G_2$ $s-t$ cut severing the t-links as per (15), will also result in severing edges in $E_2^{intra}$ (Figure 1 and (14)). In particular, $e_{im,in} \in E_2^{intra}$ will be severed iff the $s-t$ cut leaves $v_{im}$ connected to one terminal, say $s$ (resp. $t$), while $v_{in}$ remains connected to the other terminal $t$ (resp. $s$) (Figure 3). The local cost of severing intra-links in $G_2$ to assign $l_i$ to vertex $v_i$ in $G$ can be calculated as

$$D_i^{intra}(l_i) = \sum_{m=1}^{b} \sum_{n=m+1}^{b} (l_{im} \oplus l_{in}) w_{v_{im},v_{in}} \tag{16}$$

where $\oplus$ denotes binary XOR, which ensures adding the edge weight between $v_{im}$ and $v_{in}$ to the cut cost iff the cut results in one vertex connected to one terminal ($s$ or $t$) while the vertex connected to the other terminal ($t$ or $s$).

The final data term penalty is the sum of (15) and (16),

$$D_i(l_i) = D_i^{tlinks}(l_i) + D_i^{intra}(l_i). \tag{17}$$

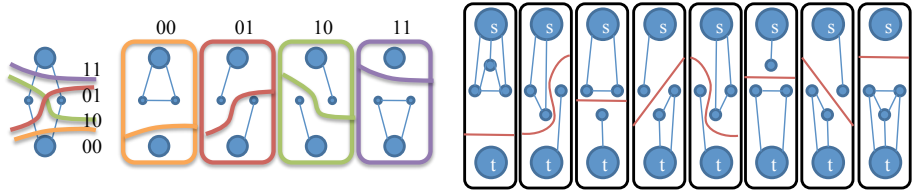

Figure 3: The $2^b$ ways to cut through $\{v_{ij}\}_{j=1}^{b}$ are shown for $b=2$ (left) and for $b=3$ (right). Note that the severed t-links and intra-links for each case follow (15) and (16), respectively.

## 2.3 Prior term penalty: Severing n-links

The vertex interaction penalty, $V_{ij}(l_i,l_j,d_i,d_j)$ in (2), for assigning $l_i$ to $v_i$ and $l_j$ to neighboring $v_j$ in $G(V,E)$, i.e. $e_{ij} \in E$, equals the cost for assigning a sequence of binary labels $(l_{im})_{m=1}^b$ to $(v_{im})_{m=1}^b$ and $(l_{jn})_{n=1}^b$ to $(v_{in})_{n=1}^b$ in $G_2(V_2, E_2)$. The local (corresponding only to labeling $v_i$ and $v_j$) cost of this cut can be calculated as (Figure 4)

$$V_{ij}(l_i,l_j,d_i,d_j) = \sum_{m=1}^b \sum_{n=1}^b (l_{im} \oplus l_{jn}) w_{v_{im},v_{jn}}. \tag{18}$$

This effectively adds the edge weight between $v_{im}$ and $v_{jn}$ to the cut cost iff the cut results in one vertex of the edge connected to one terminal ($s$ or $t$) while the other vertex connected to the other terminal ($t$ or $s$). Note that we impose no restrictions on the left hand side of (18), *e.g.* it could reflect non-convex or non-metric priors, and can be spatially varying. Essentially, for every pair $(i,j)$, $V_{ij}(l_i,l_j,d_i,d_j)$ must only return a non-negative scalar. As special cases, $V_{ij}$ could be $V_{ij}^l(l_i,l_j), V_{ij}^d(d_i,d_j)$, or $V_{ij}^l(l_i,l_j)V_{ij}^d(d_i,d_j)$.

## 2.4 Edge weight approximation with least squares

Equations (17) and (18) dictate the relationship between the penalties of the data and prior terms ($D_i$ and $V_{ij}$) of the original multi-label problem (that of $G(V,E)$) and the severed edge weights of the binary $s-t$ cut formulation ($G(V_2,E_2)$). What remains missing before applying the $s-t$ cut, however, is to find the edge weights for the binary problem, *i.e.* $w(e_{v_{ij},v_{mn}}) = w_{ij,mn}; \forall e_{ij,mn} \in E_2$.

### 2.4.1 Edge weights of t-links and intra-links

For $b=1$ (*i.e.* binary labelling), (16) simplifies to

$$D_i^{intra}(l_i) = 0 \tag{19}$$

and (15) and (17) simplify to

$$D_i(l_i) = D_i^{tlinks}(l_i) + 0 = l_{i1} w_{v_{i1},s} + \bar{l}_{i1} w_{v_{i1},t}. \tag{20}$$

With $l_i = l_{i1}$ for $b=1$, substituting the two possible values for $l_i$, $l_i = l_0$ and $l_i = l_1$, we obtain these two equations

$$\begin{aligned} l_i = l_0 &\Rightarrow D_i(l_0) = l_0 w_{v_{i1},s} + \bar{l}_0 w_{v_{i1},t} = 0 w_{v_{i1},s} + 1 w_{v_{i1},t} \\ l_i = l_1 &\Rightarrow D_i(l_1) = l_1 w_{v_{i1},s} + \bar{l}_1 w_{v_{i1},t} = 1 w_{v_{i1},s} + 0 w_{v_{i1},t} \end{aligned} \tag{21}$$

which can be written in matrix form $A_1 X_1^i = B_1^i$ as

$$\begin{pmatrix} 0 & 1 \\ 1 & 0 \end{pmatrix} \begin{pmatrix} w_{v_{i1},s} \\ w_{v_{i1},t} \end{pmatrix} = \begin{pmatrix} D_i(l_0) \\ D_i(l_1) \end{pmatrix} \tag{22}$$

where $X_1^i$ is the vector of unknown edge weights connecting vertex $v_{i1}$ to $s$ and $t$, $B_1^i$ is the data term penalty for $v_i$, and $A_1$ is the matrix of coefficients. The subscript 1 in $A_1, X_1^i$, and $B_1^i$ indicate that this matrix equation is for the case of $b = 1$. Clearly, the solution to (22) is trivial

$$w_{v_{i1},s} = D_i(l_1) \text{ and } w_{v_{i1},t} = D_i(l_0). \tag{23}$$

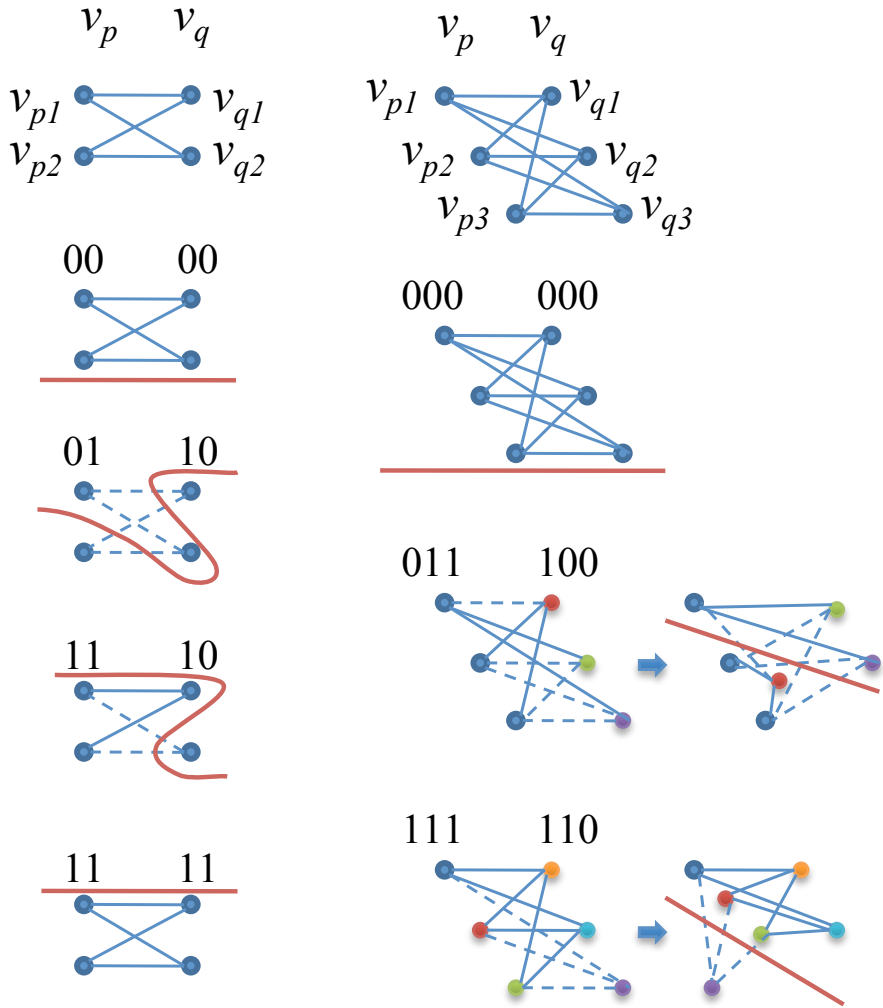

Figure 4: Severing n-links between neighboring vertices $v_p$ and $v_q$ for $b = 2$ (four such examples are shown in the left column) and $b = 3$ (3 examples in the right column). The cut is depicted as a red curve. In the last two examples for $b = 3$, the colored vertices are translated while maintaining the n-links in order to clearly show that the severed n-links for each case follow (18).

*i.e.* the higher the penalty of assigning $l_1$ to $v_i$, the more costly it is to sever $e_{v_{i1},s}$ and hence the more likely it is to assign $l_0$ to $v_i$, and vice versa. This agrees with what we expect in the binary case. The more interesting cases are when $b > 1$.

For $b = 2$, we address mutli-label problems with $2^{b-1} = 2 < k \leq 2^b = 4$ labels, *i.e.* $k = 3$ or $k = 4$. Substituting the $2^b = 4$ possible label values, $((0,0),(0,1),(1,0),$ and $(1,1))$, of $(l_i)_2 = (l_{i1}, l_{i2})$ in (17) we obtain

$$D_i(l_i) = \sum_{j=1}^{2} l_{ij} w_{v_{ij},s} + \bar{l}_{ij} w_{v_{ij},t} + \sum_{m=1}^{2} \sum_{n=m+1}^{2} (l_{im} \oplus l_{in}) w_{v_{im},v_{in}}. \tag{24}$$

$$\begin{aligned}
(0,0) &\Rightarrow D_i(l_0) = 0w_{v_{i1},s} + 1w_{v_{i1},t} + 0w_{v_{i2},s} + 1w_{v_{i2},t} + 0w_{v_{i1},v_{i2}} \\
(0,1) &\Rightarrow D_i(l_1) = 0w_{v_{i1},s} + 1w_{v_{i1},t} + 1w_{v_{i2},s} + 0w_{v_{i2},t} + 1w_{v_{i1},v_{i2}} \\
(1,0) &\Rightarrow D_i(l_2) = 1w_{v_{i1},s} + 0w_{v_{i1},t} + 0w_{v_{i2},s} + 1w_{v_{i2},t} + 1w_{v_{i1},v_{i2}} \\
(0,0) &\Rightarrow D_i(l_3) = 1w_{v_{i1},s} + 0w_{v_{i1},t} + 1w_{v_{i2},s} + 0w_{v_{i2},t} + 0w_{v_{i1},v_{i2}}
\end{aligned} \tag{25}$$

which can be written in matrix form $A_2 X_2^i = B_2^i$ as

$$\begin{pmatrix} 0 & 1 & 0 & 1 & 0 \\ 0 & 1 & 1 & 0 & 1 \\ 1 & 0 & 0 & 1 & 1 \\ 1 & 0 & 1 & 0 & 0 \end{pmatrix} \begin{pmatrix} w_{v_{i1},s} \\ w_{v_{i1},t} \\ w_{v_{i2},s} \\ w_{v_{i2},t} \\ w_{v_{i1},v_{i2}} \end{pmatrix} = \begin{pmatrix} D_i(l_0) \\ D_i(l_1) \\ D_i(l_2) \\ D_i(l_3) \end{pmatrix}. \tag{26}$$

Similarly, for $b = 3$ ($k = 5, 6, 7$, or $8$), we write $2^b = 8$ equations to the linear system of equations $A_3 X_3^i = B_3^i$, where

$$A_3 = \begin{pmatrix}
 & l_{i1}l_{i2}l_{i3} & l_{i1} & \bar{l}_{i1} & l_{i2} & \bar{l}_{i2} & l_{i3} & \bar{l}_{i3} & l_{i1} \oplus l_{i2} & l_{i1} \oplus l_{i3} & l_{i2} \oplus l_{i3} \\
000 \Rightarrow & & 0 & 1 & 0 & 1 & 0 & 1 & 0 & 0 & 0 \\
001 \Rightarrow & & 0 & 1 & 0 & 1 & 1 & 0 & 0 & 1 & 1 \\
010 \Rightarrow & & 0 & 1 & 1 & 0 & 0 & 1 & 1 & 0 & 1 \\
011 \Rightarrow & & 0 & 1 & 1 & 0 & 1 & 0 & 1 & 1 & 0 \\
100 \Rightarrow & & 1 & 0 & 0 & 1 & 0 & 1 & 1 & 1 & 0 \\
101 \Rightarrow & & 1 & 0 & 0 & 1 & 1 & 0 & 1 & 0 & 1 \\
110 \Rightarrow & & 1 & 0 & 1 & 0 & 0 & 1 & 0 & 1 & 1 \\
111 \Rightarrow & & 1 & 0 & 1 & 0 & 1 & 0 & 0 & 0 & 0
\end{pmatrix} \tag{27}$$

$$X_3^i = (w_{v_{i1},s}, w_{v_{i1},t}, w_{v_{i2},s}, w_{v_{i2},t}, w_{v_{i3},s}, \\ w_{v_{i3},t}, w_{v_{i1},v_{i2}}, w_{v_{i1},v_{i3}}, w_{v_{i2},v_{i3}})^t \tag{28}$$

$$B_3^i = (D_i(l_0), D_i(l_1), D_i(l_2), \cdots, D_i(l_7))^t. \tag{29}$$

In general, for any $b$, we have

$$A_b X_b^i = B_b^i \tag{30}$$

where $X_b^i$ is a column vector of length $2b + \binom{b}{2}$

$$\begin{aligned}X_b^i = (&w_{v_{i1},s}, w_{v_{i1},t}, w_{v_{i2},s}, w_{v_{i2},t}, \cdots, w_{v_{ib},s}, w_{v_{ib},t}, \\ &w_{v_{i1},v_{i2}}, w_{v_{i1},v_{i3}}, \cdots, w_{v_{i1},v_{ib}}, w_{v_{i2},v_{i3}}, \cdots, w_{v_{i2},v_{ib}}, \cdots, w_{v_{i,b-1},v_{ib}})^t\end{aligned} \quad (31)$$

$A_b$ is a $2^b \times (2b + \binom{b}{2})$ matrix whose $j$th row $A_b(j,:)$ is

$$\begin{aligned}A_b(dec(l_{i1}l_{i2}\cdots l_{ib}),:) = (&l_{i1}, \bar{l}_{i1}, l_{i2}, \bar{l}_{i2}, \cdots, l_{ib}, \bar{l}_{ib}, \\ &l_{i1} \oplus l_{i2}, l_{i1} \oplus l_{i3}, \cdots, l_{i1} \oplus l_{ib}, l_{i2} \oplus l_{i3}, l_{i2} \oplus l_{ib}, \cdots, l_{i,b-1} \oplus l_{ib})\end{aligned} \quad (32)$$

where $dec(.)$ is the decimal equivalent of its binary argument. $B_b^i$ is a $2^b$-long column vector given by

$$B_b^i = \left(D_i(l_0), D_i(l_1), D_i(l_2), \cdots, D_i(l_{2^b-1})\right)^t. \quad (33)$$

We can now solve the linear system of equations in (30) and find the optimal, in a LS sense, t-links and intra-links edge weights $\hat{X}_b^i$ related to every vertex $v_i$ using

$$\hat{X}_b^i = A_b^+ B_b^i \quad (34)$$

where $A^+$ is the (Moore-Penrose) pseudoinverse of $A$ calculated using singular value decomposition (SVD): If $A = U\Sigma V^*$ is the SVD of $A$ then $A^+ = V\Sigma^+ U^*$, where the diagonal elements of $\Sigma^+$ are the reciprocal of each non-zero element of $\Sigma$ [1, 34, 35].

Solving (34) for every vertex $v_i$, we obtain the weights of all edges in $E_2^{tlinks} \cup E_2^{intra}$. Note that $A_b$ and, more importantly, $A_b^+$ are easily pre-computed off-line only once for each $b$ value, as they do not change for different vertices or for different graphs.

### 2.4.2 Edge weights of n-links

For $b = 1$ (*i.e.* binary labelling), (18) simplifies to

$$(l_i \oplus l_j) w_{ij} = V_{ij}(l_i, l_j, d_i, d_j) \quad (35)$$

where $w_{v_{i1},v_{j1}}$ has been replaced by $w_{i,j}$ and $l_{i1}$ and $l_{j1}$ have been replaced by $l_i$ and $l_j$, since they are equivalent in the $b = 1$ case.

In the case when the vertex interaction depends on the data only and is independent of the labels $l_i$ and $l_j$, *i.e.* $V_{ij}(l_i, l_j, d_i, d_j) = V_{ij}^d(d_i, d_j)$, we can simply ignore the outcome of $l_i \oplus l_j$ and thus set it to a constant $1/c$, then the solution is trivial $w_{i,j} = cV_{ij}(d_i, d_j)$, which agrees with (4). However, in the general case, when $V_{ij}$ depends on the labels $l_i$ and $l_j$ of the neighboring vertices $v_i$ and $v_j$, a single edge weight is insufficient to capture such elaborate label interactions essentially because $w_{i,j}$ needs to take on a different value for every pair of labels.

To address this problem, we substitute in (18) each of the $2^b 2^b = 2^{2b} = 2^2 = 4$ possible combinations of pairs of labels $(l_i, l_j) \in \{l_0, l_1\} \times \{l_0, l_1\} = \{0, 1\} \times \{0, 1\}$, and obtain:

$$\begin{aligned}(l_0, l_0) = (0, 0) &\Rightarrow V_{ij}(l_0, l_0, d_i, d_j) = (0 \oplus 0) w_{i,j} = 0 \\ (l_0, l_1) = (0, 1) &\Rightarrow V_{ij}(l_0, l_1, d_i, d_j) = (0 \oplus 1) w_{i,j} = w_{i,j} \\ (l_1, l_0) = (1, 0) &\Rightarrow V_{ij}(l_1, l_0, d_i, d_j) = (1 \oplus 0) w_{i,j} = w_{i,j} \\ (l_1, l_1) = (1, 1) &\Rightarrow V_{ij}(l_1, l_1, d_i, d_j) = (1 \oplus 1) w_{i,j} = 0\end{aligned} \quad (36)$$

which is written in matrix form $S_1 Y_1^{ij} = T_1^{ij}$ as

$$\begin{pmatrix} 0 \\ 1 \\ 1 \\ 0 \end{pmatrix} (w_{i,j}) = \begin{pmatrix} V_{ij}(l_0, l_0, d_i, d_j) \\ V_{ij}(l_0, l_1, d_i, d_j) \\ V_{ij}(l_1, l_0, d_i, d_j) \\ V_{ij}(l_1, l_1, d_i, d_j) \end{pmatrix} \quad (37)$$

where $Y_1^{ij}$ is the unknown n-link weight $w_{i,j}$ connecting $v_i$ to neighboring $v_j$. As before, subscript 1 indicates $b = 1$. The first and fourth equations capture the condition that in order to guarantee the same label for neighboring vertices then the edge weight should be infinite ($0/V_{ij}$) and, hence, never severed. Solving for $w_{ij}$ using pseudoinverse gives $w_{ij} = S_1^+ T_1^i = \frac{1}{2}(V_{ij}(l_0, l_1, d_i, d_j) + V_{ij}(l_1, l_0, d_i, d_j))$ since $S_1^+ = (0, 0.5, 0.5, 0)$, i.e. $w_{ij}$ is equal to the average between the interaction penalties of the two cases when the labels are different.

For $b = 2$, (18) simplifies to

$$V_{ij}(l_i, l_j, d_i, d_j) = (l_{i1} \oplus l_{j1}) w_{v_{i1}, v_{j1}} + \\ (l_{i1} \oplus l_{j2}) w_{v_{i1}, v_{j2}} + (l_{i2} \oplus l_{j1}) w_{v_{i2}, v_{j1}} + (l_{i2} \oplus l_{j2}) w_{v_{i2}, v_{j2}} \quad (38)$$

We can now substitute all possible $2^b 2^b = 2^{2b} = 16$ combinations of the pairs of interacting labels $(l_i, l_j) \in \{l_0, l_1, l_2, l_3\} \times \{l_0, l_1, l_2, l_3\}$, or equivalently, $((l_i)_2, (l_j)_2) \in \{00, 01, 10, 11\} \times \{00, 01, 10, 11\}$. Here are some examples,

$$\begin{aligned}
(l_0, l_0) &= (00, 00) \Rightarrow V_{ij}(l_0, l_0, d_i, d_j) = 0 w_{v_{i1}, v_{j1}} + 0 w_{v_{i1}, v_{j2}} + 0 w_{v_{i2}, v_{j1}} + 0 w_{v_{i2}, v_{j2}} \\
(l_0, l_1) &= (00, 01) \Rightarrow V_{ij}(l_0, l_1, d_i, d_j) = 0 w_{v_{i1}, v_{j1}} + 1 w_{v_{i1}, v_{j2}} + 0 w_{v_{i2}, v_{j1}} + 1 w_{v_{i2}, v_{j2}} \\
(l_0, l_2) &= (00, 10) \Rightarrow V_{ij}(l_0, l_2, d_i, d_j) = 1 w_{v_{i1}, v_{j1}} + 0 w_{v_{i1}, v_{j2}} + 1 w_{v_{i2}, v_{j1}} + 0 w_{v_{i2}, v_{j2}} \\
(l_0, l_3) &= (00, 11) \Rightarrow V_{ij}(l_0, l_3, d_i, d_j) = 1 w_{v_{i1}, v_{j1}} + 1 w_{v_{i1}, v_{j2}} + 1 w_{v_{i2}, v_{j1}} + 1 w_{v_{i2}, v_{j2}} \\
(l_1, l_0) &= (01, 00) \Rightarrow V_{ij}(l_1, l_0, d_i, d_j) = 0 w_{v_{i1}, v_{j1}} + 0 w_{v_{i1}, v_{j2}} + 1 w_{v_{i2}, v_{j1}} + 1 w_{v_{i2}, v_{j2}} \\
(l_2, l_1) &= (10, 01) \Rightarrow V_{ij}(l_2, l_1, d_i, d_j) = 1 w_{v_{i1}, v_{j1}} + 0 w_{v_{i1}, v_{j2}} + 0 w_{v_{i2}, v_{j1}} + 1 w_{v_{i2}, v_{j2}} \\
(l_3, l_3) &= (11, 11) \Rightarrow V_{ij}(l_3, l_3, d_i, d_j) = 0 w_{v_{i1}, v_{j1}} + 0 w_{v_{i1}, v_{j2}} + 0 w_{v_{i2}, v_{j1}} + 0 w_{v_{i2}, v_{j2}}
\end{aligned} \quad (39)$$

Writing all the 16 equations, we obtain the linear system of equations in matrix format as $S_2 Y_2^{ij} = T_2^{ij}$, where $Y_2^{ij} = (w_{v_{i1}, v_{j1}}, w_{v_{i1}, v_{j2}}, w_{v_{i2}, v_{j1}}, w_{v_{i2}, v_{j2}})^t$ is the $4 \times 1$ vector of unknown n-link edge weights, $T_2^{ij}$ is a $16 \times 1$ vector whose entries are the different possible interaction penalties $((V_{ij}(l_i, l_j, d_i, d_j))_{i=0}^3)_{j=0}^3$, and $S_2$ is a $16 \times 4$ matrix with 0 or 1 entires resulting from $\oplus$ as follows

$$S_2 = \begin{pmatrix} 0 & 0 & 1 & 1 & 0 & 0 & 1 & 1 & 1 & 0 & 0 & 1 & 1 & 0 & 0 \\ 0 & 1 & 0 & 1 & 0 & 1 & 0 & 1 & 1 & 0 & 1 & 0 & 1 & 0 & 1 & 0 \\ 0 & 0 & 1 & 1 & 1 & 1 & 0 & 0 & 0 & 0 & 1 & 1 & 1 & 1 & 0 & 0 \\ 0 & 1 & 0 & 1 & 1 & 0 & 1 & 0 & 0 & 1 & 0 & 1 & 1 & 0 & 1 & 0 \end{pmatrix}^t. \quad (40)$$

In general, for any $b$, we have the following linear system of equations

$$S_b Y_b^{ij} = T_b^{ij} \quad (41)$$

where $Y_b^{ij}$ is the $b^2 \times 1$ vector of unknown n-link edge weights, $S_b$ is $2^{2b} \times b^2$ matrix of 0s and 1s, and $T_b^{ij}$ is a $2^{2b} \times 1$ vector of interaction penalties, *i.e.*

$$\begin{pmatrix} s_{0,0} \\ s_{0,1} \\ \vdots \\ s_{0,2^b-1} \\ s_{1,0} \\ s_{1,1} \\ \vdots \\ s_{1,2^b-1} \\ \vdots \\ s_{2^b-1,0} \\ s_{2^b-1,1} \\ \vdots \\ s_{2^b-1,2^b-1} \end{pmatrix} \begin{pmatrix} w_{i1,j1} \\ w_{i1,j2} \\ \vdots \\ w_{i1,jb} \\ w_{i2,j1} \\ w_{i2,j2} \\ \vdots \\ w_{i2,jb} \\ \vdots \\ w_{ib,j1} \\ w_{ib,j2} \\ \vdots \\ w_{ib,jb} \end{pmatrix} = \begin{pmatrix} V_{ij}(l_0,l_0,d_i,d_j) \\ V_{ij}(l_0,l_1,d_i,d_j) \\ \vdots \\ V_{ij}(l_0,l_{2^b-1},d_i,d_j) \\ V_{ij}(l_1,l_0,d_i,d_j) \\ V_{ij}(l_1,l_1,d_i,d_j) \\ \vdots \\ V_{ij}(l_1,l_{2^b-1},d_i,d_j) \\ \vdots \\ V_{ij}(l_{2^b-1},l_0,d_i,d_j) \\ V_{ij}(l_{2^b-1},l_1,d_i,d_j) \\ \vdots \\ V_{ij}(l_{2^b-1},l_{2^b-1},d_i,d_j) \end{pmatrix} \tag{42}$$

where $s_{m,n}$ is a row in $S_b$ and is given by

$$\begin{aligned} s_{m,n} = (&l_{m1} \oplus l_{n1}, l_{m1} \oplus l_{n2}, \cdots, l_{m1} \oplus l_{nb}, \\ &l_{m2} \oplus l_{n1}, l_{m2} \oplus l_{n2}, \cdots, l_{m2} \oplus l_{nb}, \cdots, \\ &l_{mb} \oplus l_{n1}, l_{mb} \oplus l_{n2}, \cdots, l_{mb} \oplus l_{nb}). \end{aligned} \tag{43}$$

We now solve the linear system of equations in (41) to find the optimal, in a LS sense, n-links edge weights $\hat{Y}_b^{ij}$ related to a pair of vertices $v_i$ and $v_j$ using

$$\hat{Y}_b^{ij} = S_b^+ T_b^{ij}. \tag{44}$$

Similar to what we noted for (34), $S_b$ and $S_b^+$ are pre-computed off-line only once for each $b$ value.

Solving (44) for every pair of neighboring vertices $v_i$ and $v_j$, we obtain the weights of all edges in $E_2^{inter}$, and solving (34) for every vertex $v_i$, we obtain the weights of all edges in $E_2^{tlinks} \cup E_2^{intra}$, *i.e.* $w_{ij,mn}, \forall e_{ij,mn} \in E_2$ are now known. We now calculate the minimal $s-t$ cut of $G_2$ to obtain the binary labeling of every vertex in $V_2 = \{\{v_{ij}\}_{i=1}^{|V|}\}_{j=1}^{b}$. Finally, every sequence of $b$ binary labels $(v_{ij})_{j=1}^{b}$ is decoded to a decimal label $l_i \in \mathscr{L}_k = \{l_0, l_1, ..., l_{k-1}\}, \forall v_i \in V$, *i.e.* the solution to the original multi-label MRF problem.

## 2.5 Gray encoding for extra labels

In cases when $b$ bits are needed to represent $k$ labels (according to (10)) but with $k < 2^b$, *e.g.* when $b = 2$ and $2^b = 4$ but $k = 3$, or when $b = 3$ and $2^b = 8$ but $k = 5, 6,$ or 7 (but not 8), we have what we call *extra or unused labels*: The *n*th label $l_{n-1}$ is extra iff $k < n \leq 2^b$ (remember that $\mathscr{L}_k = \{l_0, l_1, ..., l_{k-1}\}$), *e.g.* the $4^{th}$ label is an extra label when $k = 3$, the $6^{th}$, $7^{th}$ and $8^{th}$ labels are extra labels when $k = 5$, *etc.*. Following an $s-t$ cut, we can, in general, end up with these extra labels, and must, therefore, replace or merge them with any of the *non-extra or used* labels: The $m$th label $l_{m-1}$ is a non-extra label iff $2^{b-1} < m \leq k$. If label $l_n$ is an extra label to be replaced with label $l_m$, then, we must replace $D_i(l_n)$ with $D_i(l_m)$ when substituting (as in (25)) all possible label values in (17). Similarly, we must replace $V_{ij}(l_n, l_j, d_i, d_j)$ by $V_{ij}(l_m, l_j, d_i, d_j)$ and $V_{ij}(l_i, l_n, d_i, d_j)$ by $V_{ij}(l_i, l_m, d_i, d_j)$ when substituting (as in (39)) all possible combinations of the pairs of interacting labels in (18). Rather than merging arbitrary labels, we adopt a Gray encoding scheme. That is, we minimize the Hamming distance (HD)[2] between the binary codes of a pair of merged labels. For example, we favor merging label 0001 with 1001 (HD=1) over merging 0001 with 0010 (HD=2). To implement this, we first note that the most significant bit of the binary code of an extra label will always be 1 (if it isn't, then we'll be using more bits than needed). Then, each extra label is merged with the non-extra label whose binary code is identical to that of the extra label *except* for having 0 as its most significant bit. Thus guaranteeing HD=1 for all pairs of merged labels. For example, 100 will be merged with 000, 111 with 011, *etc.*, or more generally $(l_n)_2 = (1, l_2, \cdots, l_b)_2$ is merged with $(l_m)_2 = (0, l_2, \cdots, l_b)_2$.

# 3 Results

## 3.1 LS error and rank deficiency analysis

The approximation error for general LS problems is a well studied topic [23, 20, 4]. In our method, to estimate the edge weights of t-links and intra-links in (44), a system of $2^b$ linear equations are solved for $2b + \binom{b}{2}$ unknowns, compared to $2^{2b}$ equations and $b^2$ unknowns when estimating the n-links edge weights in (44). Table 1 summarizes the number of equations, unknowns, and the ranks of $A_b$ and $S_b$ of for different values of $b$. Note that the only full-rank case is $A_1$ (*i.e.* binary segmentation). $A_b$ is underdetermined for $b = 2, 3$ and overdetermined for $b \geq 4$. All cases of $S_b$ are rank deficient and overdetermined.

We present, in Figure 5, empirical results of LS error $e_b$ when solving for the edge weights of t-links and intra-links (*c.f.* Section 2.4.1, (30), (34)), and, in Figure 6, the error $e_t$ of n-links (*c.f.* Section 2.4.2 and (41), (44)). $e_b$ and $e_t$ are given by

$$e_b = |B_b^i - \hat{B}_b^i|/|B_b^i| = |(I - A_b A_b^+)B_b^i|/|B_b^i| \qquad (45)$$

$$e_t = |T_b^{ij} - \hat{T}_b^{ij}|/|T_b^{ij}| = |(I - S_b S_b^+)T_b^{ij}|/|T_b^{ij}| \qquad (46)$$

where $I$ is the identity matrix and $|.|$ is the $l^2$-norm. Note how the error in Figure 5 starts at exactly zero for binary segmentation ($b = 1$), as expected. With increasing number of labels, the average error increases with an (empirical) upper bound of 0.5, whereas the error variance decreases. In Figure 6, the error is non-zero even for binary segmentation (Section 2.4.2) and it converges to 0.5 as the number of labels increases. The plots are the result of a Monte Carlo simulation of 500 random realizations of the constant vectors $B_b^i$ and $T_b^{ij}$ (the right hand side of (30) and (41)) for each number of labels.

---

[2]The Hamming distance between two strings of equal length (two binary codes in our case) is the number of positions for which the corresponding symbols (or bits) is different.

| $b$ | $A_b$ in (30) | | | | | $S_b$ in (41) | | | | |
|---|---|---|---|---|---|---|---|---|---|---|
| bits | $e$ | $u$ | $r$ | $u_0$ | $r_0$ | $e$ | $u$ | $r$ | $u_0$ | $r_0$ |
| 1 | 2 | 2 | 2 | 2 | 2 | 4 | 1 | 1 | 1 | 1 |
| 2 | 4 | 5 | 4 | 4 | 3 | 16 | 4 | 4 | 2 | 2 |
| 3 | 8 | 9 | 7 | 6 | 4 | 64 | 9 | 9 | 3 | 3 |
| 4 | 16 | 14 | 11 | 8 | 5 | 256 | 16 | 16 | 4 | 4 |
| 5 | 32 | 20 | 16 | 10 | 6 | 1024 | 25 | 25 | 5 | 5 |
| 6 | 64 | 27 | 22 | 12 | 7 | 4096 | 36 | 36 | 6 | 6 |
| 7 | 128 | 35 | 29 | 14 | 8 | 16384 | 49 | 49 | 7 | 7 |
| 8 | 256 | 44 | 37 | 16 | 9 | 65536 | 64 | 64 | 8 | 8 |

Table 1: Properties of the system of linear equations. For different numbers of bits $b$, the table lists the number of equations $e$, number of unknowns $u$, and ranks $r$ of the matrix of coefficients $A_b$ (c.f. (30) in section 2.4.1) and $S_b$ (c.f. (41) in section 2.4.2). $u_0$ and $r_0$ reflect the case when, for $A_b$, intra-links are not used (i.e. $E_2^{intra} = \emptyset$ in (11)) and, for $S_b$, only sparse n-links are used (i.e. $E_2^{nf} = \emptyset$ in (13)).

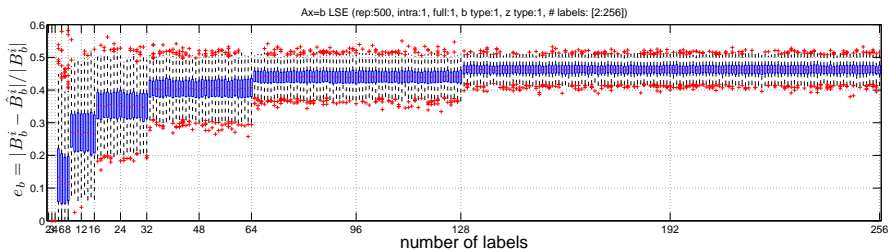

Figure 5: LS error $e_b$ in estimating the t-links and intra-links edge weights for increasing number of labels.

## 3.2 Effect of LS error in edge weights on $s-t$ cut

Our inability to model the multi-way cut exactly as an $s-t$ cut is captured by the LS error in estimating the edge weights. This error in edge weights results in error in the $s-t$ cut (or error in the binary labeling), which is then decoded into a suboptimal solution to the multi-label problem. In Figure 7 and Figure 8, we quantify the error in the cut cost and the labeling accuracy due to edge weight errors for different numbers of labels. To this end, we create a graph $G$ with a proper topology (i.e. reflecting the 4-connectedness of 2D image pixels) and random edge weights (sampled uniformly from $[0,1]$). We then construct $G_{LSE}$, a noisy version of $G$, by introducing errors in the edge weights modeled after the LS error (*i.e.* the norm of the error is dependent on the number of labels, according to the error analysis results in Figure 5 and Figure 6). The cut cost error $\triangle|C|$ is calculated using

$$\triangle|C| = ||C| - |C_{LSE}||/|C| \qquad (47)$$

where $|C| = \sum_{e_{ij} \in C} w_{ij}$ is cut cost of $G$ and $|C_{LSE}|$ is the cut cost of $G_{LSE}$. The labeling accuracy *ACC* is calculated using

$$ACC = (TP + TN)/|V| \qquad (48)$$

where $TP + TN$ gives the total number of correctly labelled, as object or background, vertices (*i.e.* true positive and true negatives), and $|V|$ is the total number of vertices in the graph, which is equal to the number of pixels in the image times the number of bits needed to encode the different labels. The plots are the results of a Monte Carlo simulation of 10 realizations of $G$ and $G_{LSE}$ representing a $25 \times 25$-pixel image, with the number of labels ranging from 2 to 256. Note that LSE errors were introduced to edge weights even for the binary case (Section 2.4.2), which explains why $ACC < 1$ and $\triangle|C| > 0$ for $b = 1$. We obtained an average (over all numbers of labels and all noise realizations) $\triangle|C| = 0.094$ and $ACC = 0.864$ with standard deviations 0.0009 and 0.0054, respectively. Note also the encouraging behavior where $\triangle|C|$ and *ACC* remain almost constant even for increasing number of labels. Increasing the number of pixels by 16 times to $100 \times 100$ and doubling the number of realizations to 20, the reported values, for 128 labels, remained almost constant with an average $\triangle|C| = 0.0926$ and $ACC = 0.863$, with standard deviation 0.00018 and 0.0011, respectively. Note, however, that in image segmentation scenario the image intensities will be corrupted by noise, in addition to the errors introduced by the LS error.

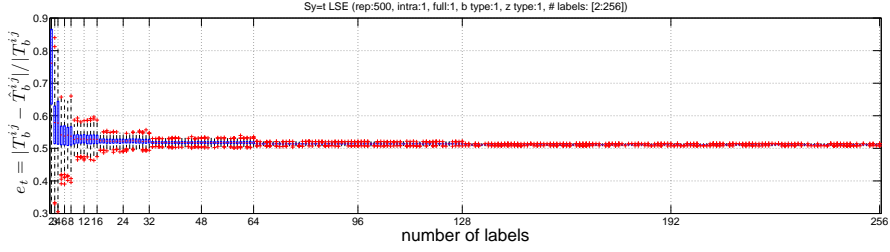

Figure 6: LS error $e_t$ in estimating the n-links for increasing number of labels.

It is important to emphasize that if we naively corrupt the edge weights of $G$ with random error rather than LS error, we will obtain different *ACC* and $\triangle|C|$ values with increasing number of labels. To show this, we create the graph $G$ as before, but now the noisy version of $G$ is created by simply adding noise sampled uniformly from $[0, \text{noise level}]$ to the edge weights of $G$. The results are given in Figure 9 and Figure 10.

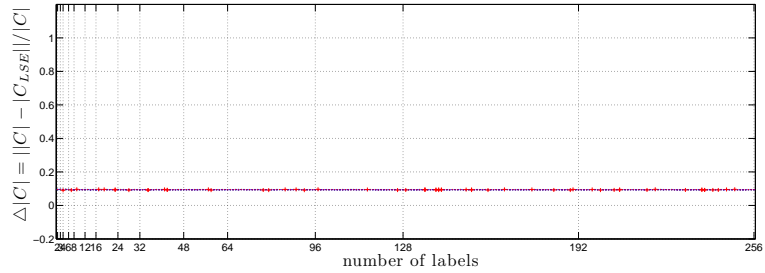

Figure 7: Cut cost error $\triangle|C|$ for increasing number of labels when the error in edge weights is induced by the LS error.

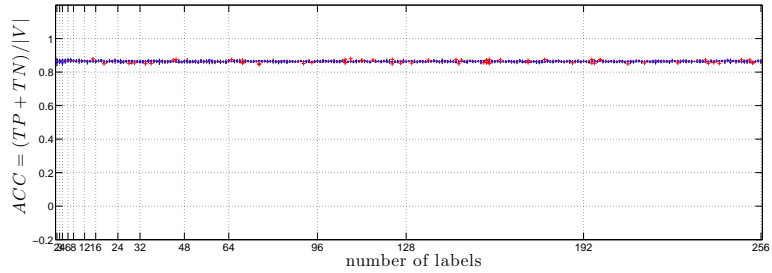

Figure 8: Labeling accuracy *ACC* for increasing number of labels when the error in edge weights is modeled after the LS error.

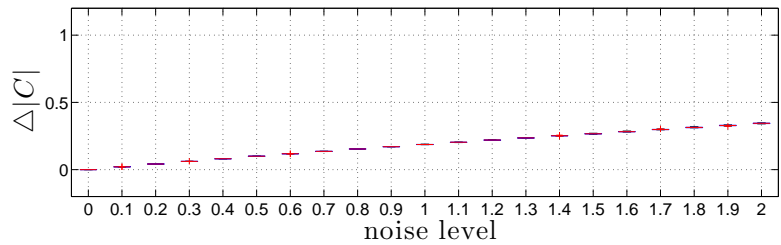

Figure 9: Cut cost error $\triangle|C|$ for increasing number of labels as we corrupt the edge weights with random (not LS) error.

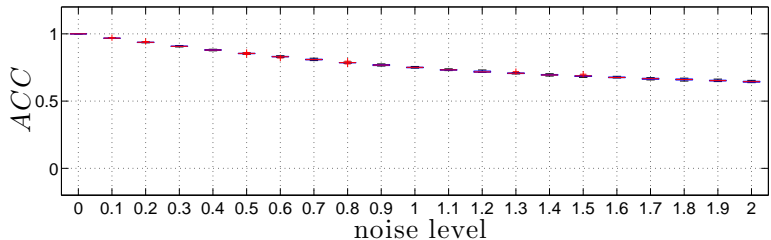

Figure 10: Labeling accuracy *ACC* for increasing number of labels as we corrupt the edge weights with random (not LS) error.

## 3.3 Image segmentation results

We evaluate our algorithm's segmentation results on synthetic images by calculating $\overline{DSC}$ (Figure 11). We tested increasing levels of additive white Gaussian noise (AWGN), with 9 standard deviation levels $\sigma \in \{0, 0.05, 0.10, \cdots, 0.40\}$, corrupting images ($I(x,y) : R^2 \to [0,1]$) of ellipses with random orientations, random lengths of major/minor axes', and varying pixel intensities. We tested 15 numbers of labels $k = \{2, 3, 4, \cdots, 16\}$ ($k-1$ ellipses plus the background label). Sample images are shown in Figure 12. We examined 11 different values for $\lambda = \{0, 0.1, 0.2, \cdots, 1\}$ (see (1)). We used the Pott's label interaction penalty ($V_{ij}^l = \delta_{l_i \neq l_j}$) with a spatially varying Gaussian image intensity penalty ($V_{ij}^d = exp(-\beta (d_i - d_j)^2)$) with $\beta = 1$ (see Section 1.2.). 50% of the pixels of each region (or label) $l$ of the noisy image (mimicking seeding) were used to learn a Gaussian probability density function $p_l(x) \sim N(\mu_l, \sigma_l)$ of the image intensity $x$ for that region. The data penalty $D_i(l_i)$ for each pixel $i$ with intensity $x_i$ was calculated as $(p_l(\mu) - p_l(x_i))/p_l(\mu)$. We ran 10 realization for each test case, *i.e.* a total of 14,850 segmentations ($9 \times 15 \times 11 \times 10$).

From Figure 11, we note high $\overline{DSC}$ for small number of labels and small noise levels and, as expected, gradually decreasing $\overline{DSC}$ results with increasing labels and noise. Note, for example, the topmost blue curve for $\sigma = 0.05$ shows almost perfect segmentation ($\overline{DSC} = 1$), whereas the second from top green curve for $\sigma = 0.1$, shows that $\overline{DSC}$ drops below 1 as the number of labels is 9 or higher. For $\sigma = 0.15$ this drop occurs earlier, at 5 labels.

We also present qualitative segmentation results on synthetic data (Figure 12) and on magnetic resonance brain images (Figure 13) from BrainWeb [11].

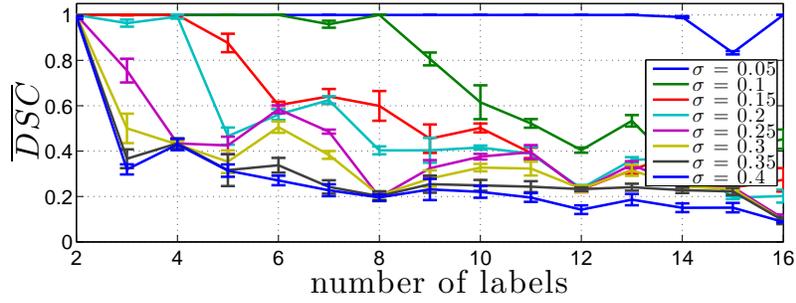

Figure 11: Dice similarity coefficient $\overline{DSC}$ between ground truth segmentation and our method's segmentation versus increasing number of labels and for noise levels (different colors).

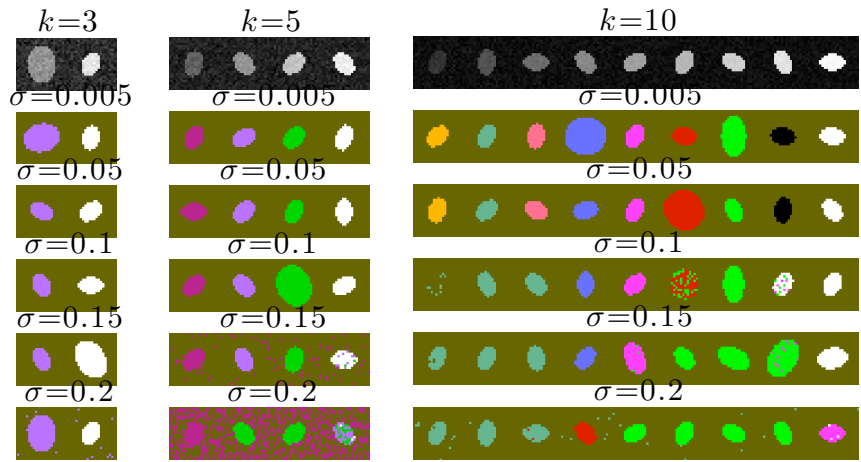

Figure 12: Sample qualitative results on images of ellipses with $k$ labels ($k-1$ ellipses plus background) and noise level $\sigma$. (top row) sample intensity images; (remaining rows) labeling results.

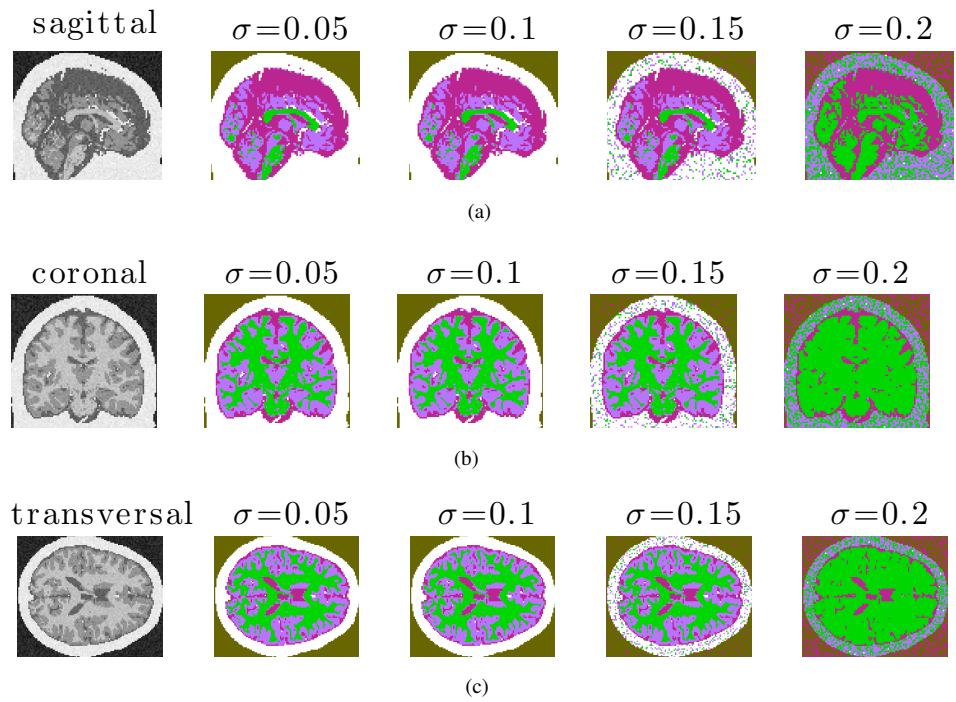

Figure 13: Brain MRI segmentation on (a) sagittal, (b) coronal and (c) transversal slices for increasing noise $\sigma$.

# 4  Conclusions

Multi-label MRF optimization is a challenging problem especially with non-trivial label-interaction priors. Algorithms that address these challenges have numerous implications for a variety of computer vision applications (*e.g.* segmentation, stereo reconstruction, *etc.*). We presented a novel approach to examining multi-label MRF. Rather than labeling a single vertex with one of *k* labels, each vertex is first replaced by $b = ceil(log_2(k))$ new vertices, and every new vertex is binary-labelled. The binary labeling of the new vertices encodes the original *k* labels, effectively approximating the multi-label problem with a globally and non-iteratively solvable $s-t$ cut. With *b* vertices replacing each original vertex, a new graph topology emerges, whose edge weights are approximated using a novel LS error approach, derived from a system of linear equation capturing the original multi-label MRF energy without any restrictions on the interaction priors. Offline pre-computation of the pseudo-inverse used in LS is performed only once and used for different graphs and vertices. We quantitatively evaluated different properties of the proposed approximation method and demonstrated the application of our approach to image segmentation (with qualitative and quantitative results on synthetic and brain images).

Future research is focused on addressing some of the deficiencies of the presented work as well as exploring ideas for improvements. The segmentation results will likely be improved with proper optimization of the free parameters (Section 1.2) (e.g. the choice of the label-interaction prior $V_{ij}^l(l_i, l_j)$, the spatially adaptive data interaction $V_{ij}^d(d_i, d_j)$, their associated parameters, $T$ and $\beta$, and $\lambda$ that balances the data and prior terms). Following such parameter optimization, it will be essential to compare with other approaches for multi-label segmentation methods. For segmentation of images that are more complex than intensity images, e.g. color images, diffusion tensor magnetic resonance images, dynamic positron tomography images, *etc.*, the data interaction term must be replaced to better capture distances between vector and tensor pixels rather than scalar pixels. We plan to evaluate the performance of the method on computer vision problems that necessitates non-metric label interaction.

We noted $\triangle|C|$ and *ACC* remaining almost constant with increasing number of labels when corrupting the graphs with LS error rather than random noise (Figure 7 and Figure 8). We speculate the reason is that the number of unknowns does not increase as fast as the number of equations, but this remains to be further investigated and formally explored.

# 5  Acknowledgements